\documentclass[runningheads]{llncs}

\usepackage{eccv}

\usepackage{eccvabbrv}

\usepackage{graphicx}
\usepackage{booktabs}

\usepackage[accsupp]{axessibility}  % Improves PDF readability for those with disabilities.

\usepackage{float}
\usepackage{times}
\usepackage{epsfig}
\usepackage{graphicx}
\usepackage{amsmath}
\usepackage{amssymb}
\usepackage{pifont}
\usepackage{colortbl}
\usepackage{multirow}
\usepackage{diagbox}
\usepackage{xcolor}
\usepackage{svg}
\usepackage{wrapfig,lipsum,booktabs}
\definecolor{iceblue}{RGB}{33,102,200}
\definecolor{fired}{RGB}{222,82,57}
\usepackage[export]{adjustbox}
\definecolor{mygray}{gray}{.9}
\definecolor{mygreen2}{RGB}{80,100,40}  % green
\definecolor{mygreen3}{RGB}{80,99,43}
\definecolor{myred}{RGB}{147,58,50}
\definecolor{myyellow}{RGB}{255,192,0}
\makeatletter
\newcommand{\thickhline}{%
    \noalign {\ifnum 0=`}\fi \hrule height 1pt
    \futurelet \reserved@a \@xhline
}
\newcommand{\reshx}[2]{
	{#1}\fontsize{7.5pt}{1em}\selectfont{\color{fired}{$\downarrow$\textbf{#2}}}
}

\usepackage[pagebackref,breaklinks,colorlinks,citecolor=eccvblue]{hyperref}
\usepackage{orcidlink}
\begin{document}

\title{AMD: Automatic Multi-step Distillation of \\ Large-scale Vision Models}

\titlerunning{Automatic Multi-step Distillation}

\author{Cheng Han\inst{1,2}\orcidlink{0000-0002-8145-3436} \and
Qifan Wang\inst{3}
% \orcidlink{1111-2222-3333-4444}
\and
Sohail A. Dianat \inst{2} \and
Majid Rabbani \inst{2} \and
Raghuveer M. Rao \inst{4} \and
Yi Fang \inst{5} \and
Qiang Guan\inst{6} \and
Lifu Huang\inst{7} \and
% Dongfang Liu\inst{2}\footnote{Corresponding author.}\orcidlink{1111-2222-3333-4444}
Dongfang Liu$^{\star}$\inst{2}
% \footnote{Corresponding author.}
% \footnote{Corresponding author.}
% \footnote{Corresponding author.}
% Qiang Guan\inst{2,3}\orcidlink{1111-2222-3333-4444} \and
}

\authorrunning{C. Han, Q, Wang, et al.}

\institute{University of Missouri -- Kansas City \and
Rochester Institute of Technology \and
Meta AI \and
DEVCOM Army Research Laboratory \and
Santa Clara University \and
Kent State University \and
Virginia Tech
% \\
% \email{chk9k@umsystem.edu, wqfcr@fb.com, sadeee@rit.edu, mxreee@rit.edu, raghuveer.m.rao.civ@army.mil, yfang@scu.edu, qguan@kent.edu, lifuh@vt.edu, dongfang.liu@rit.edu}
}

% \maketitle

% Define the affiliations (institutes)
% \makeatletter
% \newcommand\inst[1]{\unskip$^{#1}$}
% \makeatother

% Sample text
\maketitle
% This is a sample document.

% Institutes (Example)
\renewcommand{\thefootnote}{\fnsymbol{footnote}}
\footnotetext[1]{Corresponding author.}
\renewcommand{\thefootnote}{\arabic{footnote}}

\begin{abstract}
% With the size of the vision foundational models continues to scale up, model compression becomes extremely important when considering current limited computation resources on mobile devices. Knowledge distillation is famous by guiding the learning of a small student network from a teacher. While it is already being a common practice in convolutional neural network, its application in transformer-based architectures remain unexplored. In this paper, we find that traditional knowledge distillation methods suffer from low accuracy
%Recent advancements in foundation models have significantly reshaped the landscape of visual recognition.
Transformer-based architectures have become the de-facto standard models for diverse vision tasks owing to their superior performance. As the size of the models continues to scale up, model distillation becomes extremely important in various real applications, particularly on devices limited by computational resources. However, prevailing knowledge distillation methods exhibit diminished efficacy when confronted with a large capacity gap between the teacher and the student, e.g, 10$\times$ compression rate.
%most famous term of foundation models, have emerged superior, overshadowing traditional convolutional neural networks (CNNs) in terms of performance. However, this superiority comes at the cost of increased computational complexity. As these models expand in size to further amplify their capabilities, they inadvertently pose significant deployment impediments, especially on devices constrained by computational resources.
%Knowledge distillation, designed to mitigate such computational demands, has shown promise in CNNs. Yet, as models continue to scale, a discernible performance deficit emerges, underscoring an imperative need for research tailored towards large-scale model distillation.
In this paper, we present a novel approach named Automatic Multi-step Distillation (AMD) for large-scale vision model compression. In particular, our distillation process unfolds across multiple steps. Initially, the teacher undergoes distillation to form an intermediate teacher-assistant model, which is subsequently distilled further to the student. An efficient and effective optimization framework is introduced to automatically identify the optimal teacher-assistant that leads to the maximal student performance.
%In this paper, we navigate the challenge of leveraging these transformer-based models in resource-constrained settings, and propose an effective yet efficient approach for large vision model compression.
%executed in a single trial, with a dual focus: optimizing training schedules for large-scale model distillation and ensuring peak performance efficacy.
% via knowledge distillation in a single trail, considering both the efficiency of training schedules on large-scale model distillation and efficacy of performance.
% Empirical results demonstrate that our approach outperform several baselines on multiple image classification datasets (\ie, CIFAR-10, CIFAR-100, ImageNet).
We conduct extensive experiments on multiple image classification datasets, including CIFAR-10, CIFAR-100, and ImageNet. The findings consistently reveal that our approach outperforms several established baselines, paving a path for future knowledge distillation methods on large-scale vision models.
%We conduct comprehensive experiments across multiple image classification datasets (\ie, CIFAR-10, CIFAR-100, ImageNet). The results indicate that our method consistently surpasses several established baselines, paving a path for future knowledge distillation methods on large-scale vision models.
% Knowledge distillation emerges as a solution, aiming to transfer knowledge from large models to compact ones. However, conventional distillation, based on the teacher-student paradigm, has shown performance declines when there's a significant capacity gap between the models. An alternative, employing teacher assistants for knowledge distillation, has gained traction. The efficacy of this approach largely depends on the optimal selection of the teacher assistant, with existing strategies leaning towards exhaustive grid search mechanisms.
\end{abstract}

\section{Introduction}
\label{sec:intro}
% why model compression is important
Foundation models in vision (\eg, BiT~\cite{kolesnikov2020big}, ViT~\cite{dosovitskiy2020image}, Swin~\cite{liu2021swin}, Florence~\cite{yuan2021florence}) have recently captured significant attention due to their revolutionary performance across various tasks. Notably, transformer-based architectures such as ViT-Large~\cite{dosovitskiy2020image} (61.6 G FLOPS) and Swin-Large~\cite{liu2021swin} (103.9 G FLOPS), which represent a general class of visual foundation models~\cite{han20232vpt, wangvisual, hanprototypical, hanfacing}, have achieved unprecedented success, demonstrating considerably more intricate operations compared to Convolutional Neural Networks (CNNs) such as ResNet18~\cite{he2016deep} (1.8 G FLOPS). However, as model scales continue to grow to enhance performance, their bottleneck in inherent high complexity presents challenges for deployment on low-power processors and mobile devices, often characterized by constrained computational capabilities~\cite{kim2023imf, li2022efficientformer, howard2017mobilenets, sandler2018mobilenetv2}.

\begin{figure}[t!]
  \centering
       % \vspace{-16pt}
       \hspace{-1.5em}
\includegraphics[width=0.90\textwidth]{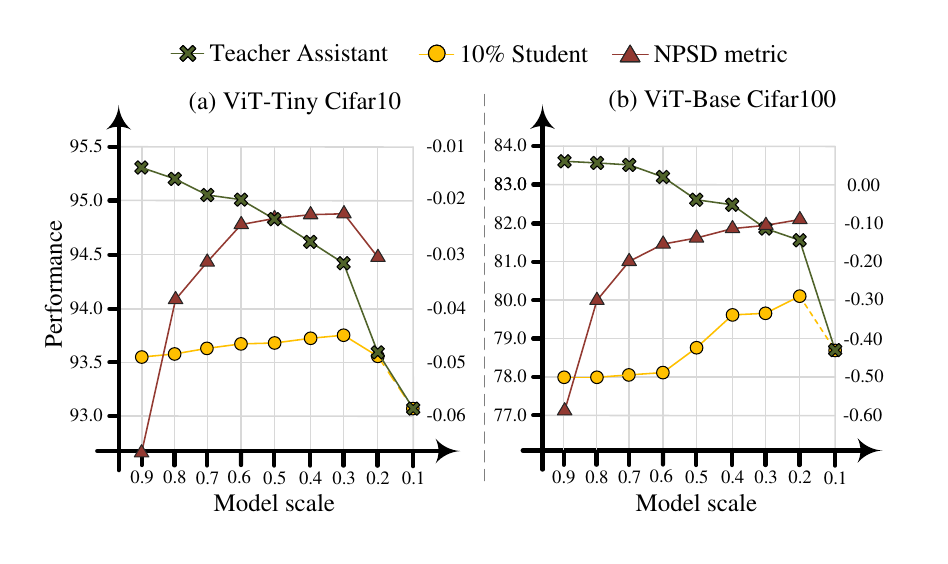}
\vspace{-6mm}
\caption{\textbf{A preliminary study on the impact of teacher-assistants with different scales and performance \textit{w.r.t.} the performance of the student.} In (a) and (b), ViT-Tiny and ViT-Base are used as the teacher models, and distilled to a 10\% student via teacher-assistants~\cite{mirzadeh2020improved} at different scales on CIFAR-10 and CIFAR-100~\cite{krizhevsky2009learning} respectively. There are several key observations: \textbf{\ding{182}} The performance of the teacher-assistant degrades when its scale decreases (see \textcolor{mygreen3}{green} curve); \textbf{\ding{183}} The performance of the student varies with different teacher-assistants in scales (see \textcolor{myyellow}{yellow} curve); \textbf{\ding{184}} We ascertain that the Negative Performance-Scale Derivative (NPSD) metric (see \S\ref{subsec:s_p_tradeoff}) exhibits a positive correlation with the performance of student models (see \textcolor{myred}{red} curve).}
\label{fig:performance_curve}
\vspace{-7mm}
\end{figure}

% what is knowledge distillation, what is the limitation of traditional multi-step distillation methods, what is the limiation of multistep distillation
A conventional approach to harnessing the superior performance of large-scale models while balancing limited computational resources is through knowledge distillation \cite{hinton2015distilling}. This process transfers the learned knowledge of these large teacher models to compact and deployable student models. Despite their effectiveness in various applications, recent studies \cite{mirzadeh2020improved, WuRGH021, cao2023learning} have revealed that traditional distillation methods suffer from severe performance degradation when a substantial capacity gap exists between the teacher and student models. Consequently, multi-step distillation approaches \cite{WangW0B0020, son2021densely} have been proposed, where the teacher is first distilled to an intermediate teacher-assistant model. The teacher-assistant then serves as an alternative teacher to better transfer knowledge to the student.
While multi-step distillation generally improves student performance, our preliminary study (Figure \ref{fig:performance_curve}) illustrates that the choice of the teacher-assistant significantly impacts the student's performance. However, due to the lack of guidance on the selection of scale/size of the teacher-assistant, trial training, namely traversing all possible scales, is usually required to search for the optimal teacher-assistant, incurring high computational costs.

To this demand, we propose an \underline{A}utomatic \underline{M}ulti-step \underline{D}istillation (AMD) approach for large vision model compression.
AMD operates through a cascade strategy comprising three phases. \textit{Structural Pruning:} a grating and pruning algorithm is employed to generate a series of teacher-assistant architectures across varying scales; \textit{Joint Optimization:} a joint optimization framework is developed to efficiently obtain the best teacher-assistants at all scales in a single run; \textit{Optimal Selection:} the optimal teacher-assistant is selected from all candidates based on the Negative Performance-Scale Derivative (NPSD) metric (see \S\ref{subsec:s_p_tradeoff}). This metric is designed to measure the performance-scale optimality of the candidates.
% Specifically, we define a criteria function $\mathbb{C}_{ta}$, depending only on the performance of teacher assistant and teacher (see Figure~\ref{}) such that the optimal teacher assistant can be targeted without going through a complete knowledge distillation trail to the student.
% This design significantly conserves computational resources during training while still reaping the performance benefits afforded by the inclusion of a teacher-assistant.
% % Such the design can greatly save the computational overhead during training while still enjoy the performance gain provided by teacher assistant.
% % In order to obtain the optimal teacher assistant according to xxx,
% Based on this design,
% OMD is implemented in a cascade strategy, which is further divided into three phases: \textbf{\ding{182}} \textit{Sparsification:} grating and pruning approach is employed to generate a series of teacher assistant candidates across varying scales; \textbf{\ding{183}} \textit{Optimization:} we further show in \S\ref{subsec:st_optimization} that the generated candidates in \textit{Sparsification} phase enjoy the xxx; \textbf{\ding{184}} \textit{Preference:} the optimal teacher assistant is selected by the smallest value by $\mathbb{C}_{ta}$ across different scales.

AMD is an efficient and powerful approach for large-scale vision model knowledge distillation. We experimentally show that, with transformer-based architectures ViT~\cite{dosovitskiy2020image} and Swin~\cite{liu2021swin}, AMD outperforms concurrent single-step and multi-step knowledge distillation methods in terms of \texttt{top-1} accuracy, \ie, \textbf{1.79-15.91\%} on CIFAR-10~\cite{krizhevsky2009learning}, \textbf{1.81-24.25\%} on CIFAR-100~\cite{krizhevsky2009learning} and \textbf{2.77-4.03\%} on ImageNet~\cite{ImageNet}. More impressively, it achieves much faster training speed (e.g., 10$\times$) compared with other multi-step distillation baselines. Comprehensive ablative studies on AMD with student model scalability, single teacher-assistant efficiency, training schedule, sampling rate and loss components further demonstrate its robustness and effectiveness. We also extend AMD to traditional CNN-based architectures in the supplementary material following common practices~\cite{zhao2022decoupled, son2021densely, shu2021channel, he2019knowledge, tian2022adaptive}, showing competitive performance to state-of-the art methods.
% , highlighting the sufficiency of using single teacher assistant,
% showing that our proposed method enjoys even lower training schedule
% These results are particularly impressive, considering the rare studies on large-scale vision model and the low computational cost of OMD. We feel this work brings fundamental insights into related fields.
%Overall, the results presented herein are especially noteworthy given the paucity of research on large-scale vision models, coupled with the minimal computational expenditure associated with our proposed AMD. It is our conviction that this work contributes fundamental understanding to the related domains.
In conclusion, the presented results are particularly significant considering the limited research on vision model compression with large gaps. We believe that this work contributes to the foundational understanding of related domains.

\begin{figure*}[t!]
  \centering
       % \vspace{-16pt}
\resizebox{0.99\textwidth}{!}{
\includegraphics[width=0.9\textwidth]{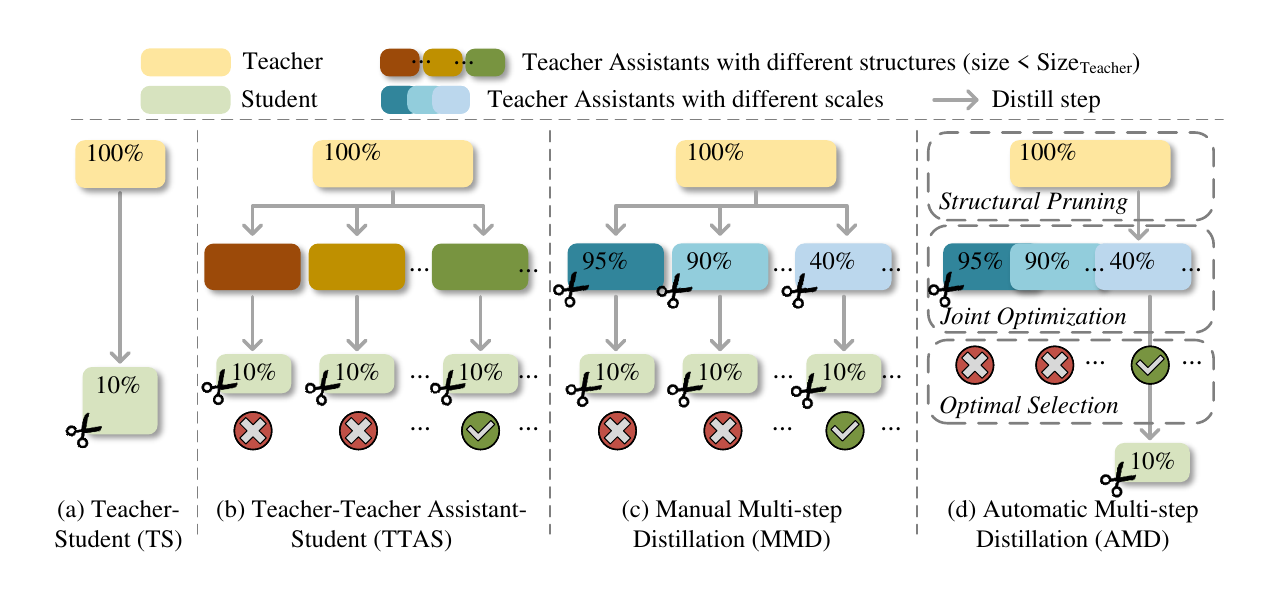}
%\includesvg[inkscapelatex=false]{imgs/architecture_omd_svg.svg}
}
% \vspace{-20pt}
\vspace{-5pt}
\caption{Overview of different knowledge distillation approaches.
(a) Traditional distillation methods directly distill the teacher to the student. (b) Multi-step distillation methods first distill the teacher to a teacher-assistant (requires a large search), which is then further distilled to the student. (c) Manual Multi-step Distillation (MMD) effectively identifies a set of teacher-assistants with different scales and performs multi-step distillation. (d) Automatic Multi-step Distillation (AMD) efficiently and effectively selects the optimal teacher-assistant through one single optimization, including three stages: \textit{Structural Pruning}, \textit{Joint Optimization} and \textit{Optimal Selection}.}
% (see \S\ref{subsec:st_optimization}).
% \textbf{The overall pipeline of our proposed optimal multi-step distillation for large vision model compression (OMD).}
% In (a) for ${\rm{OMD}_{Max}}$, one have to get the performance from each teacher assistant to search for the optimal one for student training. In (b) for ${\rm{OMD}_{Min}}$, on the other hand, schedules the optimal teacher assistant through one single trail, which is mindful for computational efficiency. This compact design relies on the cascade strategy, which include three stages: \textit{sparsification}, \textit{optimization} and \textit{preference} (see \S\ref{subsec:st_optimization}).
\label{fig:architecture}
\vspace{-5mm}
\end{figure*}

\section{Related Work}
\label{sec:related_work}
\vspace{-2mm}
\paragraph{Large-scale Vision Models}
\label{subsec:vision_transformer}
% Transformer, everything scale up
Drawing upon the notable advancements of transformers in natural language processing (NLP)~\cite{brown2020language,devlin2018bert,liu2019roberta, raffel2020exploring,vaswani2017attention}, researchers have subsequently developed the transformer architecture into various vision tasks, including image classification~\cite{dosovitskiy2020image,liu2022swin,liu2021swin, wangvisual}, image segmentation~\cite{strudel2021segmenter,wang2021max, wang2021end,zheng2021rethinking, ni2023dual}, object detection~\cite{beal2020toward,carion2020end,pan20213d,yuan2021temporal,zhu2020deformable, zhang2023structured, zhixing2021distilling}, \etc.
% Self-supervised pretraining paradigms~\cite{bao2021beit, chen2021empirical, he2022masked} has also been explored, leading to state-of-the-art results.
Transformers dominate in visual-related disciplines due to their superior performance and scalability compared to convolutional neural networks (CNNs)~\cite{he2022parameter, jia2022visual}. However, the significant complexity  overhead on transformers~\cite{fournier2021practical, islam2022recent, zhang2022patchformer, han20232vpt, yan2023prompt} also hinders their deployments on real-world scenarios with limited resources. For example, famous transformer-based architecture ViT-Base~\cite{dosovitskiy2020image} cost 2.23 times computational overhead to ResNet101~\cite{he2016deep} (\ie, 17.6G FLOPS $vs$ 7.9G FLOPS).
Despite the desire to harness the superior capabilities of large-scale vision models, prevailing research has predominantly centered on knowledge distillation within traditional convolutional neural networks, neglecting the burgeoning necessity for scale reduction in transformer-based architectures.
% While we still want to enjoy the strong performance these large-scale vision models brings, current research mostly focus on knowledge distillation on convolutional neural networks, ignoring the fact that transformer-based archiectures are in a even more emergency need for smaller scalability.
In the light of this view, we propose AMD, a knowledge distillation approach to reduce the computational overhead of large-scale vision models while maintaining satisfying performance. \\
% For instance, recent transformer-based models such as MViTv2-Large~\cite{li2021improved} (218M), ViT-G~\cite{zhai2022scaling} (1.8B), SwinV2-G~\cite{liu2022swin} (3.0B), and V-MoE~\cite{riquelme2021scaling} (14.7B) incur substantial computational costs. Therefore, we propose ${\rm E^{2}VPT}$, which is designed to reduce the computational cost of transformer-based architectures while maintaining high performance in the \textit{pretrain-then-finetune} paradigm.
% \vspace{-5mm}
% \paragraph{Model Pruning}
\noindent \textit{Model Pruning}
% \label{subsec:Model_Pruning}
Model pruning can be generally categorized into unstructured~\cite{park2020lookahead, guo2016dynamic, dong2017learning, han20232vpt} and structured pruning~\cite{ding2019centripetal, li2016pruning, you2019gate, fang2023depgraph, ma2023llm, beyer2022knowledge}. The fundamental distinction between the two approaches is that structural pruning entails a modification of the neural network architecture through the physical removal of grouped parameters, whereas unstructured pruning involves zeroing out selected weights without altering the intrinsic structure of the network~\cite{fang2023depgraph}. Though unstructured pruning enjoys a finer-grained pruning, it fits only to specialized devices. Structural pruning, on the other hand, does not necessitate specialized accelerators or software frameworks to achieve reductions in memory utilization and computational overhead~\cite{Minidisc2023, wu2023ad}. Consequently, it is more amenable to a broader range of practical applications. In transformer-based architecture, a series of works have been presented to identify a design structure in desired scale (\eg, dynamic search~\cite{hou2020dynabert}, layer dropping~\cite{fan2019reducing} and pruning~\cite{michel2019sixteen}).
In our work, we employ structured pruning for deriving the structures of candidates for its beneﬁts for distillation. \\
% Additionally, pruning offers an opportunity to optimize the efﬁciency of our method due to its merits.
% \vspace{-5mm}
% \paragraph{Knowledge Distillation}
\noindent \textit{Knowledge Distillation}
% \label{subsec:KD}
Knowledge distillation~\cite{hinton2015distilling} has drawn much attention in vision community in the field of model compression~\cite{son2021densely, han2015deep, yu2018nisp}. The basic concept is to compress the knowledge of a larger scale model (\ie, teacher) to a computational efficient smaller one (\ie, student). Extensive research has been conducted in recent years, which can
% Knowledge distillation can
be broadly categorized into two groups: single-step and multi-step distillation. Single-step methods are first inspired by the design of~\cite{hinton2015distilling} and introduce the teacher's predictions as ``soft'' labels~\cite{kim2017transferring, muller2019does}, which directly mimic the final prediction of the final prediction from the teacher model~\cite{gou2021knowledge}.
% , and develop research on the effectiveness of the soft targets by label smoothing~\cite{kim2017transferring} and regularizers~\cite{muller2019does}.
Besides the logit-based methods~\cite{li2023curriculum, cho2019efficacy, zhao2022decoupled}, feature-mimicking methods~\cite{bengio2013representation, romero2014fitnets, heo2019knowledge, passban2021alp, chen2021cross, wang2020exclusivity} are applied in various studies to leverage the spatial-wise knowledge for intermediate features directly or indirectly \cite{huang2017like,kim2018paraphrasing,zagoruyko2016paying}.
% (\ie, derive via neuron selectivity transfer~\cite{huang2017like}, factors~\cite{kim2018paraphrasing}, attention map~\cite{zagoruyko2016paying}, \etc).
There are also relation-base studies further explore the relationships between different layers or data samples~\cite{you2017learning, park2019relational}, using knowledge from FSP matrix~\cite{yim2017gift}, instance relation~\cite{yu2019learning, peng2019correlation, chen2020learning}, similarity matrix~\cite{tung2019similarity}, mutual information flow~\cite{passalis2020heterogeneous}, \etc.
Multi-step approaches, on the other hand, emerge prominently when researchers notice that a significant model capacity discrepancy between large-scale networks and a small student can degrade knowledge transfer~\cite{mirzadeh2020improved, gao2021residual, cao2023learning}.
% \textcolor{red}{Some approaches~\cite{cao2023learning} , while most of them focus on }
Nonetheless, determining an optimal teacher-assistant for the student model presents a nuanced challenge. Methods such as \cite{son2021densely} iteratively guides multiple teacher-assistant to every other small teacher-assistants and introduces ``random drop'' for teacher or teacher-assistants to improve efficiency; \cite{jin2019knowledge} constructs a gradually mimicking sequence during learning. While these methods mitigate performance degradation during distillation, they have not gained widespread acceptance due to the elevated computational overhead associated with the teacher-assistant, especially with the increasing teacher model capacities.
% \textcolor{red}{\cite{cao2023learning} discusses the learning capacities and provides a valid solution on object detection via multi-teacher progressive distillation, while their solution on mitigate such challenge remains naive, and no further evidence are suggested on image classification. }
% \textcolor{red}{
% delves into
The learning capacities within object detection is covered by \cite{cao2023learning}, offering a viable strategy through multi-teacher progressive distillation. However, their approach to addressing this challenge appears to be in its nascent stages (see \S\ref{subsec:st_optimization}),
overlooking the potential for further narrowing the capacity gap
% ignoring the fact that the capacity gap between models can further be narrow down
via model pruning with sharing parameters.
% Moreover, the paper does not extend its findings or propose evidence regarding its applicability or efficacy in the context of image classification, leaving a gap in the literature that warrants further investigation.
% }
Furthermore, current methods~\cite{yang2022vitkd, chen2022dearkd, yao2022distill} rarely take the large-scale vision models into consideration, which often necessitate a higher compression ratio due to the considerable magnitude of these models relative to contemporary convolutional architectures.
Recent research (\eg, CSKD~\cite{zhao2023cumulative}, DeiT~\cite{touvron2021training}, CviT~\cite{ren2022co}, DearKD~\cite{chen2022dearkd}) achieves commendable performance. However, their focus is on enhancing the efficacy of large-scale models through knowledge distillation, overlooking the necessity of model compression and deployment on devices with computational constraints. In contrast, our approach is mindful of both computational efficiency and large magnitude of compression ratio, while preserving superior performance from teacher models.

\section{Methodology}\label{sec:Methodology}

% In this section, we first introduce the Performance-Scale Derivative (PSD) metric used to measure the optimality of the intermediate teacher-assistant model. Our preliminary study illustrates the effectiveness
% of PSD as the teacher-assistant measurement. We then formally define the distillation problems and notations, followed by the presentation of the proposed automatic multi-step knowledge distillation approach.
In this section, we first formally define the multi-step distillation problems and notations. Subsequently, we introduce the Negative Performance-Scale Derivative (NPSD) metric, which serves as the measure for assessing the optimality of the intermediate teacher-assistant model. Our preliminary study demonstrates the effectiveness of NPSD as the metric for evaluating teacher-assistant performance. The proposed automatic multi-step knowledge distillation approach is then presented.

\subsection{Problem Formulation}\label{subsec:problem}

% Within the Teacher-Teaching Assistant-Student paradigm (TTAS), given a teacher model $T({s_{t}}, {p_{t}})$ with scale ${s_{t}}=100\%$ and performance ${p_{t}}$ to distill into a student $S({s_{s}}, {p_{s}})$, the goal is to find an appropriate teacher assistant $TA({s_{ta}}, {p_{ta}})$ such that the student performance can be maximized.

Within the multi-step distillation paradigm (as shown in Figure \ref{fig:architecture}), the principal objective is to identify an optimal teaching-assistant, denoted as $TA(S_{ta}, P_{ta})$, which will facilitate the distillation of knowledge from a fully-scaled teacher model, $T(S_{t}, P_{t})$ where $S_{t} = 100\%$, to a student model, $S(S_{s}, P_{s})$. Here $P$ and $S$ represent the performance and scale of the models respectively.
This process aims to enhance the student model's performance, $P_{s}$, by leveraging the appropriate teaching assistant (\ie, $T \rightarrow TA\rightarrow S$). In practice, the scale $S_{ta}$ and performance $P_{ta}$ of the teacher-assistant are confined by two boundaries: for scalability, $S_{s} < S_{ta} < S_{t}$ and for performance, $P_{s} < P_{ta} < P_{t}$.

%OMD, a novel optimal multi-step knowledge distillation approach for large-scale vision model. We first define the distillation problems and notations in \S\ref{subsec:Preliminary}. The scale-performance relation is in detailed discussed in \S\ref{subsec:s_p_tradeoff}, followed by the single-trail optimization with different teacher assistant model scales in \S\ref{subsec:st_optimization}.
\subsection{Negative Performance-Scale Derivative}\label{subsec:s_p_tradeoff}
In this work, we propose to identify a suitable teacher-assistant measurement for multi-step distillation, considering the trade-off between performance and scale. Determining an appropriate objective function to optimize these factors simultaneously is non-trivial. Ideally, we aim to find a teacher-assistant with maximum performance and minimum scale. However, this ideal scenario is often not practical. On one hand, as the scale of the teacher-assistant approaches that of the student model, its performance tends to decline, making it challenging to transfer complete knowledge to the student. On the other hand, a larger scale teacher-assistant might still exhibit a significant scale gap with the student model, resulting in a substantial performance drop during the distillation process. Therefore, it is crucial to identify a suitable teacher-assistant that strikes an optimal balance between performance and scale.

To tackle this challenge, we introduce the Negative Performance-Scale Derivative (NPSD) as the optimization objective. NPSD assesses the benefits, measuring how closely the teacher-assistant can match the scale of the student, against the costs, considering the associated performance drop. A higher NPSD indicates a more favorable benefit-cost ratio. Our preliminary studies confirm that NPSD positively correlates with downstream student performances across all tasks and backbones (refer to Figure \ref{fig:performance_curve}). Based on this insight, our objective is to maximize NPSD within a manageable time complexity.
% \vspace{-4mm}
\paragraph{Definition of NPSD.}
Formally, NPSD represents the negative derivation of performance to scale, which is defined as:
\begin{equation}
\label{eq:tradeoff}
% \small
{NPSD}_{ta} = - \ \frac{P_{t} - P_{ta}}{S_{t} - S_{ta}}
\vspace{-3pt}
\end{equation}
where $t$ and $ta$ denote the teacher and teacher-assistant, respectively. $P_t$ and $P_{ta}$ represent the performances of the two models, while $S_t$ and $S_{ta}$ denote their corresponding model scales. It is essential to note that, given the teacher model, $P_t$ and $S_t$ are constants in this context.
Intuitively, a teacher-assistant with high performance and a small scale results in a high NPSD value.
% \textcolor{red}{
While NPSD calculation does not explicitly involve the student, the teach assistant's performance (\ie, $P_{ta}$) is influenced by the student model, as it is achieved through joint optimization with the student. In other words, different student scales will yield different optimal solutions for a teacher-assistant of the same size. Thus, the NPSD value of a teacher-assistant is theoretically constrained by the student.
% \textbf{\textit{Second}},
We further conduct experiments using a variation of the $\lambda$-NPSD metric, \ie, $ - (\ \frac{P_{t} - P_{ta}}{S_{t} - S_{ta}} + \lambda \cdot \frac{P_{ta} - P_{s}}{S_{ta} - S_{s}})$, explicitly considering the student model (see Table~\ref{table:npsd_metric}).
% Results with different $\lambda$ values are presented in the left table.
It is evident that $\lambda$-NPSD does not lead to better performance compared to the original NPSD from Eq.~\ref{eq:tradeoff}, but with additional complexity (\ie, a hyper-parameter $\lambda$) plus an extra distillation step in order to obtain the student's performance.
% \vspace{-4mm}
\paragraph{Preliminary Results.}
To assess the effectiveness of NPSD as the optimality measurement, we conducted multi-step distillation experiments on CIFAR-10 and CIFAR-100 \cite{krizhevsky2009learning} using various teacher models. These teacher models underwent distillation to a 10\% student through teacher-assistants~\cite{mirzadeh2020improved} at different scales ranging from \{10\%, 20\%, \dots, 80\%, 90\%\}. The architecture of each teacher-assistant model is derived through structural pruning~\cite{michel2019sixteen, wu2023ad}, which prunes the least important parameters based on their importance scores from the teacher model.

The results for ViT-Tiny on CIFAR-10 and ViT-Base on CIFAR-100 are presented in Figure \ref{fig:performance_curve}. Notably, as the scale of the teacher-assistant approaches that of the student, its performance exhibits a monotonic decrease. Conversely, the performance of students distilled from different teacher-assistants varies with the scales of the teacher-assistants.
Furthermore, we observe a positive correlation between the NPSD measure and the performance of the student models. The students achieve their highest performance when distilled with maximum NPSD values, indicating the effectiveness of NPSD measurement.
% \textcolor{red}{
Further results detailed in \S\ref{subsec:main_results} reinforce our observations.

\subsection{Automatic Multi-step Distillation}\label{subsec:st_optimization}

The objective stated above thus turns into an optimization problem of finding the optimal teacher-assistant with the maximum value of NPSD$_{ta}$.
% Formally,
% \begin{equation}
% \label{eq:argmax}
% % \small
% % \mathbb{C}_{ta} = - \frac{p_{t} - p_{ta}}{s_{t} - s_{ta}},
% \hat{TA}(\hat{s_{ta}}, \hat{p_{ta}}) = -\mathrm{argmax}~\mathbb{C}_{ta},
% \vspace{-3pt}
% \end{equation}
% where we aim to find the optimal teacher assistant $\hat{TA}(\hat{s_{ta}}, \hat{p_{ta}})$.
However, identifying an optimal teacher-assistant poses several key challenges. First, the number of potential teacher-assistants is infinitely large, given the continuous search space. Even for a fixed-scale model, its architecture can exhibit significant variation. Second, the computational cost of finding the maximum performance teacher-assistants at all scales becomes extremely high when conducting multi-step distillation for each teacher-assistant. Last but not least, one needs to select the best teacher-assistant among all candidates.

To address these challenges, we decompose the problem into three stages.
% \textcolor{red}{The first two stages together decompose the infinite searching of teacher-assistants into a finite set, where the teacher-assistants at all scales are optimized in a single trail. The third stage selects the optimal teacher-assistant.}
In \textit{Structural Pruning} stage, a series of candidate teacher-assistants at different scales is generated with gridding and pruning techniques. Then a \textit{Joint Optimization} framework is introduced to obtain the maximum performance teacher-assistants at all scales in a single optimization.
% \textcolor{red}{
The first two stages together decompose the infinite searching of teacher-assistants into a finite set, where the teacher-assistants at all scales are optimized in a single trail.
Finally, the optimal teacher-assistant with the highest NPSD is selected in the \textit{Optimal Selection} stage.
% \vspace{-5mm}

\noindent \textit{Structural Pruning.} We employ \textit{gridding} and \textit{pruning} to construct the teacher-assistant at different scales. Specifically, \textit{gridding} is first introduced to allow limited number of candidates, where we convert a continuous search space into a discrete one. Ideally, one needs to generate candidates in every possible scale in order to find a ``perfect'' teacher-assistant continuously. However, it is both impossible (\ie, subdivide infinitely) and inefficient (\ie, optimize numerous teacher-assistants jointly) to search all possible candidates in a continuous space. We thus evenly divide the scales into $m$ parts, where the gap between each candidate is $\delta = \frac{S_{t} - S_{s}}{m}$, resulting in $m$ candidates between the teacher and student. Next, to obtain the architecture of each candidate at different scales, various \textit{pruning} techniques can be applied. In this work, we adopt the structural pruning method~\cite{michel2019sixteen, wu2023ad} due to its known advantages in knowledge distillation \cite{xia2022structured}. It gradually prunes the network with the least important parameters according to their important scores, which are obtained by introducing learnable mask variables during teacher's inference stage. The resulting teacher-assistant candidates are represented as \{$\mathcal{M}_i$ $\mid$ $i$ = 1 to $m$\}.
More details are provided in the supplementary material.
% We provide more details on pruning in supplementary material.
% There are approaches such as layer dropping~
% \vspace{-5mm}
%\paragraph{Joint Optimization.}

\noindent \textit{Joint Optimization.}
The intuitive approach to finding the best teacher-assistant involves maximizing the performance of each individual candidate and subsequently employing NPSD to measure the optimality of each candidate. However, the computational overhead remains significant when assessing the performance of all candidates, given the large number of candidates. To further mitigate computational complexity, we explore parameter sharing among teacher-assistant candidates across different scales, leveraging the \textit{incremental property} inherent in candidates obtained from structural pruning. Essentially, the incremental property posits that for two candidates $\mathcal{M}_i$ and $\mathcal{M}_j$, if $S_i < S_j$, then the parameters of $\mathcal{M}_i$ form a subset of $\mathcal{M}_j$. This ensures that a larger candidate can be transformed into a smaller one by continuously pruning less significant parameters, facilitating parameter sharing across the candidates.

With parameter sharing, the total number of parameters to be optimized becomes equivalent to that of the largest candidate. Subsequently, we devise a joint optimization framework that simultaneously optimizes distillations from the teacher to all candidates in a single run. In doing so, we effectively reduce the memory footprints of all candidates to that of a distinguished one. The computational costs are also significantly diminished through the parameter-sharing optimization.
% \textcolor{red}{More importantly, we provide a natural solution to the problem mentioned in \cite{cao2023learning}, where the researchers find out that a smaller capacity gap between the student and teacher may facility knowledge transfer. While  current both homogeneous (\eg, similar architectures with different numbers of layers) and heterogeneous (\eg, different network designs) distillation need to initialize and teach the student from scratch, we propose \textit{Joint Optimization} and inherit students' parameters directly from teachers under the same architectural design, which provides a solid and effective path for student-teacher capacity gap.}
% \textcolor{red}{
More importantly, we propose a natural resolution to the issue highlighted in \cite{cao2023learning}, wherein the researchers discovered that a reduced capacity disparity between the student and teacher models can facilitate more effective knowledge transfer. Unlike existing approaches that necessitate the initialization and teaching of a separate student model from scratch, both in homogeneous distillation (\eg, similar architectures with varying numbers of layers) and heterogeneous distillation (\eg, distinct network designs), our method is \textit{orthogonal} and introduces \textit{Joint Optimization}, enabling the direct inheritance of parameters by student models from their teacher counterparts within the same, singular architectural design. This approach establishes an efficacious mechanism to bridge the capacity gap between student and teacher models, leading to the optimal distillation performance.
% , thereby enhancing the efficiency of knowledge transfer.
To sum up,
the overall objective of AMD includes the cross-entropy loss, logit-based loss, and feature-based loss:
\begin{equation}
\label{eq:overall_objective}
% \small
\mathcal{L}=\sum_{i=1}^m(\mathcal{L}_{ce}(\mathcal{T},\mathcal{M}_i)+\alpha\mathcal{L}_{logit}(\mathcal{T},\mathcal{M}_i)+\beta\mathcal{L}_{feat}(\mathcal{T},\mathcal{M}_i))
\vspace{-3pt}
\end{equation}
$\mathcal{L}_{ce}$=$CE(y^T, y^M)$ is the cross-entropy loss,
% between the output logits of the teacher $l^{t}$ and the groud-truth label $y$,
$\mathcal{L}_{logit}$=$\mathrm{KL}[\mathrm{softmax}(\frac{l^{t}}{\gamma})||\mathrm{Softmax}(\frac{l^{s}}{\gamma})]$ is the kullback-leibler divergence loss~\cite{kim2021comparing, van2014renyi} between the softened output logits, and
% between the output logits of the teacher $l^{t}$ and the student $l^{s}$.
$\mathcal{L}_{feat}=\mathrm{MSE}(H^{T}, H^{M})$ is the mean squared error between the last layer of hidden states. $y,~l,~\gamma,~H$ represent the labels, output logits, temperature value, and last layer of hidden states, respectively.
% In the context of two-stage distillation, it is noteworthy that the roles of the teacher and student are dynamic. Specifically, when delineating the relationship between the teacher and the teaching assistant, the student assumes the role of the teaching assistant. Conversely, in the relationship between the teaching assistant and the student, the teaching assistant adopts the role of the teacher. Furthermore, in scenarios involving multiple stages of teaching assistants, the loss also considers among the teaching assistants.
% \vspace{-4mm}

\noindent \textit{Optimal Selection.} The optimal teacher-assistant is then selected with the highest NPSD value from all the candidates. Subsequently, an additional distillation is taken place between selected teacher-assistant and student following the same training objective in Eq.~\ref{eq:overall_objective}. It should be acknowledged that within the domain of NLP, there exists a body of research~\cite{LiuTF022,wu2023ad} delineate the training objective through the empirical analysis of training curves in conjunction with hyper-parameter adjustments. In contrast, our method focus on the relation of teacher and teacher-assistant, facilitating the exploration of robustness across a diverse array of datasets and models, thereby contributing to a nuanced understanding of knowledge distillation dynamics.

\section{Experiment}\label{sec:Exp}
\vspace{-0.3em}

\vspace{-0.3em}
\subsection{Experimental Setup}\label{subsec:experiment_setup}
\vspace{-0.3em}
\noindent \textbf{Datasets.} The evaluation is carried out on CIFAR-10~\cite{krizhevsky2009learning}, CIFAR-100~\cite{krizhevsky2009learning} and ImageNet~\cite{ImageNet} datasets, respectively.
CIFAR-10 contains $60$K ($50$K/$10$K for train/test) colored images of $10$ classes. CIFAR-100 dataset contains $100$ classes with $500$
training and $100$ testing images per class. ImageNet contains $1.2$M images for train and $50$K images for validation of 1K classes. Note that for all experiments in transformer-based architectures, the images are in 224 $\times$ 224 resolution. In supplementary material, when considering CNN-based architectures, we instead apply resolution in 32 $\times$ 32 on CIFAR-10 and CIFAR-100 for generality.

\begin{table}[H]
% \vspace{-0.5cm}
\vspace{-3mm}
\caption{The results of knowledge distillation methods on CIFAR-10~\cite{krizhevsky2009learning} and CIFAR-100~\cite{krizhevsky2009learning} datasets. All results are reported using the same teacher and student (10\% scale). The best results are highlighted in \textbf{bold}, and the second best are shown in \underline{underline}. Follow~\cite{kim2023imf},
% \textcolor{red}{
as our student design via \textit{Structural Pruning} is \textit{orthogonal} to current approaches (see \S\ref{subsec:st_optimization}), we rerun and reproduce the results of all methods based on author-provided or author-verified code, and report the average performance over five runs. The GPU hours are estimated with respect to their conventional counterparts. \textcolor{fired}{$\downarrow$} denotes the performance lost with respect to their teacher models, a smaller gap represent a superior performance.} %Same for Table~\ref{table:main_results_imagenet}.}
\vspace{-2mm}
\label{table:main_results}
% \begin{center}
% \begin{small}
%\tabcolsep=0.10cm
% \resizebox{0.90\textwidth}{!}{
\begin{adjustbox}{width=0.87\width,center}
\begin{tabular}{l||r|c|c|c}
\hline \thickhline
% Method & FLOPs & ImageNet
\rowcolor{mygray}
Method & FLOPs & CIFAR-10~\cite{krizhevsky2009learning} \texttt{top-1} & CIFAR-100~\cite{krizhevsky2009learning} \texttt{top-1} & GPU hours \\
\hline \hline
$\rm{ViT-Tiny}_{100\%}$ (teacher) & 1.3G & 95.98\% & 76.12\%  & - \\
${\rm{ViT-Tiny}_{10\%}}$ KD \textcolor{lightgray}{\scriptsize{[arXiv15]}}~\cite{hinton2015distilling} & 0.1G & \reshx{78.75\%}{17.23\%} $\pm$ 0.32 & \reshx{48.85\%}{27.27\%} $\pm$ 0.21 & $1\times$ \\
${\rm{ViT-Tiny}_{10\%}}$ DKD \textcolor{lightgray}{\scriptsize{[CVPR22]}}~\cite{zhao2022decoupled} & 0.1G & \reshx{80.72\%}{15.26\%} $\pm$ 0.37 & \reshx{62.21\%}{13.91\%} $\pm$ 0.30 & $1\times$ \\
${\rm{ViT-Tiny}_{10\%}}$ CRD \textcolor{lightgray}{\scriptsize{[ICLR20]}}~\cite{tian2019contrastive} & 0.1G & \reshx{91.94\%}{4.04\%} $\pm$ 0.35 & \reshx{70.42\%}{5.70\%} $\pm$ 0.33 & $1\times$\\
${\rm{ViT-Tiny}_{10\%}}$ ADKD \textcolor{lightgray}{\scriptsize{[ACL23]}}~\cite{wu2023ad} & 0.1G & \reshx{84.29\%}{11.69} $\pm$ 0.42 & \reshx{66.43\%}{9.69} $\pm$ 0.37 & $1\times$\\
${\rm{ViT-Tiny}_{10\%}}$ TAKD \textcolor{lightgray}{\scriptsize{[AAAI20]}}~\cite{mirzadeh2020improved} & 0.1G & \reshx{81.57\%}{14.41\%} $\pm$ 0.67 & \reshx{62.89\%}{13.23\%} $\pm$ 0.54 & $20\times$ \\
${\rm{ViT-Tiny}_{10\%}}$ DGKD \textcolor{lightgray}{\scriptsize{[ICCV21]}}~\cite{son2021densely} & 0.1G & \reshx{88.13\%}{7.85} $\pm$ 0.60 & \reshx{69.38\%}{6.74} $\pm$ 0.52 & $84\times$\\
\rowcolor{mygray}
% ~~~~~~~~ -- w/ maximum schedule (${\rm{AMD}_{Max}}$)
${\rm{ViT-Tiny}_{10\%}}$ MMD
& 0.1G & \reshx{\underline{93.76\%}}{2.22\%} $\pm$ 0.53 & \reshx{\textbf{73.12\%}}{3.00\%} $\pm$ 0.47 & $20\times$\\
\rowcolor{mygray}
% ~~~~~~~~ -- w/ minimum schedule (${\rm{AMD}_{Min}}$)
${\rm{ViT-Tiny}_{10\%}}$ AMD
& 0.1G & \reshx{\textbf{93.83\%}}{2.15\%} $\pm$ 0.32 & \reshx{\underline{73.10\%}}{3.02\%} $\pm$ 0.30 & $2.2\times$ \\
\hline \hline
$\rm{ViT-Small}_{100\%}$ (teacher) & 4.6G & 97.69\% & 87.82\% & -  \\
${\rm{ViT-Small}_{10\%}}$ KD \textcolor{lightgray}{\scriptsize{[arXiv15]}}~\cite{hinton2015distilling} & 0.5G & \reshx{79.21\%}{18.48\%} $\pm$ 0.35 & \reshx{57.38\%}{30.44\%} $\pm$ 0.33 & $1\times$ \\
${\rm{ViT-Small}_{10\%}}$ DKD \textcolor{lightgray}{\scriptsize{[CVPR22]}}~\cite{zhao2022decoupled} & 0.5G & \reshx{85.17\%}{12.52\%} $\pm$ 0.37 & \reshx{64.42\%}{23.40\%} $\pm$ 0.34 & $1\times$ \\
${\rm{ViT-Small}_{10\%}}$ CRD \textcolor{lightgray}{\scriptsize{[ICLR20]}}~\cite{tian2019contrastive} & 0.5G & \reshx{93.12\%}{4.57\%} $\pm$ 0.36 & \reshx{77.36\%}{10.46\%} $\pm$ 0.34 & $1\times$\\
${\rm{ViT-Small}_{10\%}}$ ADKD \textcolor{lightgray}{\scriptsize{[ACL23]}}~\cite{wu2023ad} & 0.5G & \reshx{88.71\%}{9.56} $\pm$ 0.42 & \reshx{72.84\%}{14.98} $\pm$ 0.39 & $1\times$\\
${\rm{ViT-Small}_{10\%}}$ TAKD \textcolor{lightgray}{\scriptsize{[AAAI20]}}~\cite{mirzadeh2020improved} & 0.5G & \reshx{86.10\%}{11.59\%} $\pm$ 0.69 & \reshx{64.98\%}{22.84\%} $\pm$ 0.63 & $20\times$ \\
${\rm{ViT-Small}_{10\%}}$ DGKD \textcolor{lightgray}{\scriptsize{[ICCV21]}}~\cite{son2021densely} & 0.5G & \reshx{91.85\%}{5.84} $\pm$ 0.56 & \reshx{76.57\%}{11.25} $\pm$ 0.48 & $84\times$\\
\rowcolor{mygray}
% ~~~~~~~~ -- w/ maximum schedule (${\rm{AMD}_{Max}}$)
${\rm{ViT-Small}_{10\%}}$ MMD
& 0.5G & \reshx{\textbf{95.14\%}}{2.56\%} $\pm$ 0.57 & \reshx{\textbf{79.22\%}}{8.60\%} $\pm$ 0.52 & $20\times$\\
\rowcolor{mygray}
% ~~~~~~~~ -- w/ minimum schedule (${\rm{AMD}_{Min}}$)
${\rm{ViT-Small}_{10\%}}$ AMD
& 0.5G &  \reshx{\underline{95.12\%}}{2.57\%} $\pm$ 0.34 & \reshx{\underline{79.17\%}}{8.65\%} $\pm$ 0.33 & $2.2\times$ \\
\hline \hline
$\rm{ViT-Base}_{100\%}$ (teacher) & 17.6G & 98.01\%  & 89.33\% & - \\
${\rm{ViT-Base}_{10\%}}$ KD \textcolor{lightgray}{\scriptsize{[arXiv15]}}~\cite{hinton2015distilling} & 1.8G & \reshx{79.63\%}{18.38\%} $\pm$ 0.38 & \reshx{59.47\%}{29.86\%} $\pm$ 0.37 & $1\times$\\
${\rm{ViT-Base}_{10\%}}$ DKD \textcolor{lightgray}{\scriptsize{[CVPR22]}}~\cite{zhao2022decoupled} & 1.8G & \reshx{85.71\%}{12.30\%} $\pm$ 0.39 & \reshx{69.53\%}{19.80\%} $\pm$ 0.36 & $1\times$\\
${\rm{ViT-Base}_{10\%}}$ CRD \textcolor{lightgray}{\scriptsize{[ICLR20]}}~\cite{tian2019contrastive} & 1.8G & \reshx{94.18\%}{3.83\%} $\pm$ 0.41 & \reshx{78.29\%}{11.04\%} $\pm$ 0.39 & $1\times$\\
${\rm{ViT-Base}_{10\%}}$ ADKD \textcolor{lightgray}{\scriptsize{[ACL23]}}~\cite{wu2023ad} & 1.8G & \reshx{88.45\%}{9.56} $\pm$ 0.35 & \reshx{73.69\%}{15.64} $\pm$ 0.38 & $1\times$\\
${\rm{ViT-Base}_{10\%}}$ TAKD \textcolor{lightgray}{\scriptsize{[AAAI20]}}~\cite{mirzadeh2020improved} & 1.8G & \reshx{87.43\%}{10.58\%} $\pm$ 0.68 & \reshx{70.89\%}{18.44\%} $\pm$ 0.64 & $20\times$\\
${\rm{ViT-Base}_{10\%}}$ DGKD \textcolor{lightgray}{\scriptsize{[ICCV21]}}~\cite{son2021densely} & 1.8G & \reshx{92.78\%}{5.23} $\pm$ 0.55 & \reshx{78.33\%}{11.00} $\pm$ 0.53 & $84\times$\\
\rowcolor{mygray}
% ~~~~~~~~ -- w/ maximum schedule (${\rm{AMD}_{Max}}$)
${\rm{ViT-Base}_{10\%}}$ MMD
& 1.8G & \reshx{\textbf{95.54\%}}{2.47\%} $\pm$ 0.54 & \reshx{\underline{80.11\%}}{9.22\%} $\pm$ 0.53 & $20\times$\\
\rowcolor{mygray}
% ~~~~~~~~ -- w/ minimum schedule (${\rm{AMD}_{Min}}$)
${\rm{ViT-Base}_{10\%}}$ AMD
& 1.8G & \reshx{\underline{95.52\%}}{2.49\%} $\pm$ 0.38 & \reshx{\textbf{80.19\%}}{9.14\%} $\pm$ 0.37 & $2.2\times$\\
\hline \hline
$\rm{Swin-Base}_{100\%}$ (teacher) & 15.1G & 98.80\% & 92.68\% & - \\
${\rm{Swin-Base}_{10\%}}$ KD \textcolor{lightgray}{\scriptsize{[arXiv15]}}~\cite{hinton2015distilling} & 1.5G & \reshx{80.29\%}{18.51\%} $\pm$ 0.37 & \reshx{61.05\%}{31.63\%} $\pm$ 0.37 & $1\times$ \\
${\rm{Swin-Base}_{10\%}}$ DKD \textcolor{lightgray}{\scriptsize{[CVPR22]}}~\cite{zhao2022decoupled} & 1.5G & \reshx{86.16\%}{12.64\%} $\pm$ 0.41 & \reshx{72.13\%}{20.55\%} $\pm$ 0.32 & $1\times$ \\
${\rm{Swin-Base}_{10\%}}$ CRD \textcolor{lightgray}{\scriptsize{[ICLR20]}}~\cite{tian2019contrastive} & 1.5G & \reshx{94.23\%}{4.57\%} $\pm$ 0.39 & \reshx{79.71\%}{12.97\%} $\pm$ 0.32 & $1\times$\\
${\rm{Swin-Base}_{10\%}}$ ADKD \textcolor{lightgray}{\scriptsize{[ACL23]}}~\cite{wu2023ad} & 1.5G & \reshx{89.58\%}{9.22} $\pm$ 0.44 & \reshx{76.64\%}{16.04} $\pm$ 0.37 & $1\times$\\
${\rm{Swin-Base}_{10\%}}$ TAKD \textcolor{lightgray}{\scriptsize{[AAAI20]}}~\cite{mirzadeh2020improved} & 1.5G & \reshx{86.72\%}{12.08\%} $\pm$ 0.59 & \reshx{73.86\%}{18.82\%} $\pm$ 0.61 & $20\times$ \\
${\rm{Swin-Base}_{10\%}}$ DGKD \textcolor{lightgray}{\scriptsize{[ICCV21]}}~\cite{son2021densely} & 1.5G & \reshx{93.54\%}{5.26} $\pm$ 0.58 & \reshx{79.15\%}{13.53} $\pm$ 0.46 & $84\times$\\
\rowcolor{mygray}
% ~~~~~~~~ -- w/ maximum schedule (${\rm{AMD}_{Max}}$)
${\rm{Swin-Base}_{10\%}}$ MMD
& 1.5G & \reshx{\textbf{96.07}\%}{2.73\%} $\pm$ 0.50 & \reshx{\textbf{83.61}\%}{9.07\%} $\pm$ 0.47 & $20\times$\\ % \reshx{\textbf{96.04\%}}{2.76\%}
\rowcolor{mygray}
% ~~~~~~~~ -- w/ minimum schedule (${\rm{AMD}_{Min}}$)
${\rm{Swin-Base}_{10\%}}$ AMD
& 1.5G & \reshx{\underline{96.02}\%}{2.78\%} $\pm$ 0.31 & \reshx{\underline{83.55}\%}{9.13\%} $\pm$ 0.28 & $2.2\times$ \\ % \reshx{\underline{95.98\%}}{2.82\%}
\hline
\end{tabular}
\end{adjustbox}
% \end{small}
% \end{center}
\vspace{-1em}
\end{table}

\noindent \textbf{Networks and Baselines.} In this paper, our primary focus is on large-scale vision models, and we conduct experiments using various networks, namely Vision Transformer (ViT)\cite{dosovitskiy2020image} and Swin Transformer (Swin)\cite{liu2021swin} with different model sizes (ViT-Tiny, ViT-Small, ViT-Base, Swin-Base). Current vision methods predominantly center around CNN-based distillation (see \S\ref{sec:related_work}), posing challenges in transferring their specifically designed methods into Transformer-based architectures. In the context of large-scale vision models, our research reproduces and compares our proposed AMD with several widely adopted knowledge distillation methods. These include four single-step baselines: KD~\cite{hinton2015distilling}, CRD~\cite{tian2019contrastive}, DKD~\cite{zhao2022decoupled}, and ADKD~\cite{wu2023ad}, as well as two multi-step baselines: TAKD~\cite{mirzadeh2020improved} and DGKD~\cite{son2021densely}. For both multi-step methods, we use the same set of teacher-assistants as employed in AMD. In the case of DGKD, we conduct experiments with two intermediate teacher-assistants. For completeness, we further design extensive studies on CNN-based architectures in the supplementary material.

\noindent \textbf{Implementation Details.} The implementation settings of AMD follow common practices~\cite{tian2019contrastive, son2021densely, zhao2022decoupled} with preprocessing, optimization, training plan and learning rate, \etc. We use standardized preprocessing~\cite{dosovitskiy2020image} including normalization and rescaling \textit{without} applying other preprocessing techniques. We set 0.003 as the initial learning rate, and use AdamW optimizer~\cite{loshchilov2017decoupled} scheduled by a cosine schedule with 3 warm up epochs.
The training epoch is set to 160 with batch size 128, consistent to current research~\cite{son2021densely, mirzadeh2020improved}.
We set the scale sampling rate $m$ to 9, which balances the performance and computational cost (see Table~\ref{table:candidate_sampling}), resulting candidate scales $\left\{10\%, 20\%, ... , 80\%, 90\%\right\}$ after structural pruning stage. For hyper-parameter $\alpha$ and $\beta$ in Eq.~\ref{eq:overall_objective}, we follow common practice~\cite{wu2023ad}, and tune $\alpha$ from $\left\{0.0, 0.1, 0.2, 0.5, 0.7\right\}$ and $\beta$ from $\left\{1, 10, 50, 100\right\}$ respectively. After exhaustively studying $\alpha$ and $\beta$ in \S\ref{subsec:ablation}, we choose $\alpha$=0.2, $\beta$=100 for all experiment setups. We use default temperature $\gamma$=1.
For other single-step methods compared in our study, we follow~\cite{zhao2022decoupled, hinton2015distilling} and apply their default training epochs for fair comparisons (\ie, trained 240 epochs on CIFAR-10 and CIFAR-100, 100/120 epochs on ImageNet).
We utilize the publicly accessible codebases, making modifications solely to the teacher and student networks by incorporating transformer-based architectures.
% We adopt their publicly available codebase and alter only the teacher and student network into transformer-based architectures.
During training, we notice that stochastic gradient descent (SGD) with step learning rate schedule often lead to low accuracy for transformers. We thus employ the AdamW optimizer to ensure stable training.
% More implementation details are shown in supplementary material.
% In light of observations that their \textit{default} optimizer, Stochastic Gradient Descent (SGD), frequently results in diminished accuracy for transformers (refer to Table~\ref{}), we have transitioned to employing the AdamW optimizer to ensure stable training
% using stochastic gradient descent (SGD) optimizer).

\noindent \textbf{Reproducibility.} AMD is implemented in Pytorch~\cite{NEURIPS2019_9015}. Experiments are conducted on NVIDIA A100-40GB GPUs.
% Our code is available at \url{https://github.com/ChengHan111/AMD}.
Our implementation will be released for reproducibility.

\vspace{-0.3em}
\subsection{Main Results}\label{subsec:main_results}
\vspace{-0.3em}
%We respectively examine the performance and robustness of AMD on several datasets with different network architectures at different scales.
% For completeness, we provide additional results on CNN-based architectures following common practices~\cite{hinton2015distilling, zhao2022decoupled, son2021densely, mirzadeh2020improved} in supplementary material.

\noindent \textbf{Results on CIFAR-10 and CIFAR-100.}
Table~\ref{table:main_results} reports the \texttt{top-1} accuracy scores on CIFAR-10 and CIFAR-100 datasets for five runs. Several key observations are taken from these results. \textbf{First}, AMD is able to outperform current knowledge distillation methods with large performance gap in both CIFAR-10 and CIFAR-100 datasets. For example, our model achieves \textbf{1.34-15.89\%} improvement in \texttt{top-1} accuracy on ViT-Base knowledge distillation on CIFAR-10; \textbf{Second}, based on the suboptimal performance exhibited by logit-based methods (\eg, KD, DKD and TAKD) across diverse scales and datasets, it can be reasonably inferred that these methods may not represent the ideal solution for knowledge distillation in transformer-based architectures. In contrast, through the utilization of feature-space supervision, CRD, ADKD and our proposed AMD achieve commendable performance. This trend suggests a necessity for feature-mimicking studies in the context of knowledge distillation, particularly as the depth and width of the transformer-based architectures expand, in order to preserve the integrity and efficacy of the distilled performances; \textbf{Third}, our AMD achieves a much faster training speed compared with the multi-step methods, indicating the effectiveness of the structural pruning and joint optimization; \textbf{Finally}, our approach consistently outperforms other knowledge distillation methods under various transformer-based architectural design, showing a promising avenue for future model deployment.

\setlength\intextsep{10pt}
\begin{wraptable}{r}{0.65\linewidth}
% \begin{table}[t]
\vspace{-0.7cm}
\caption{Comparison results on ImageNet~\cite{ImageNet} dataset.}
% \vspace{-3mm}
\label{table:main_results_imagenet}
\resizebox{0.63\textwidth}{!}{
\begin{tabular}{l||c}
\hline \thickhline
\rowcolor{mygray}
Method & ImageNet~\cite{ImageNet} \texttt{top-1} \\
\hline \hline
$\rm{ViT-Base}_{100\%}$ (teacher) & 77.90\% \\
% ${\rm{ViT-Base}_{10\%}}$ KD \textcolor{lightgray}{\scriptsize{[arXiv]}}~\cite{hinton2015distilling} &   \\
% ${\rm{ViT-Base}_{10\%}}$ DKD \textcolor{lightgray}{\scriptsize{[CVPR22]}}~\cite{zhao2022decoupled} &  \\
% ${\rm{ViT-Base}_{10\%}}$ TAKD \textcolor{lightgray}{\scriptsize{[AAAI20]}}~\cite{mirzadeh2020improved} &  \\
${\rm{ViT-Base}_{10\%}}$ CRD \textcolor{lightgray}{\scriptsize{[ICLR20]}}~\cite{tian2019contrastive} & \reshx{69.66\%}{8.24\%} $\pm$ 0.31 \\
${\rm{ViT-Base}_{10\%}}$ ADKD \textcolor{lightgray}{\scriptsize{[ACL23]}}~\cite{wu2023ad} & \reshx{68.40\%}{9.50\%} $\pm$ 0.39 \\
${\rm{ViT-Base}_{10\%}}$ DGKD \textcolor{lightgray}{\scriptsize{[ICCV21]}}~\cite{son2021densely} & \reshx{69.25\%}{8.65\%} $\pm$ 0.44 \\
\rowcolor{mygray}
${\rm{ViT-Base}_{10\%}}$ MMD
& \reshx{\textbf{72.43}\%}{5.47\%} $\pm$ 0.27 \\
\rowcolor{mygray}
% ~~~~~~~~ -- w/ minimum schedule (${\rm{AMD}_{Min}}$)
${\rm{ViT-Base}_{10\%}}$ AMD
& \reshx{\underline{72.27}\%}{5.63\%} $\pm$ 0.22 \\
\hline
$\rm{Swin-Base}_{100\%}$ (teacher) & 83.50\% \\
% ${\rm{Swin-Base}_{10\%}}$ KD \textcolor{lightgray}{\scriptsize{[arXiv]}}~\cite{hinton2015distilling} & \\
% ${\rm{Swin-Base}_{10\%}}$ DKD \textcolor{lightgray}{\scriptsize{[CVPR22]}}~\cite{zhao2022decoupled} & \\
% ${\rm{Swin-Base}_{10\%}}$ TAKD \textcolor{lightgray}{\scriptsize{[AAAI20]}}~\cite{mirzadeh2020improved} & \\
${\rm{Swin-Base}_{10\%}}$ CRD \textcolor{lightgray}{\scriptsize{[ICLR20]}}~\cite{tian2019contrastive} & \reshx{74.83\%}{8.67\%} $\pm$ 0.28 \\
${\rm{Swin-Base}_{10\%}}$ ADKD \textcolor{lightgray}{\scriptsize{[ACL23]}}~\cite{wu2023ad} & \reshx{72.38\%}{11.12\%} $\pm$ 0.32 \\
${\rm{Swin-Base}_{10\%}}$ DGKD \textcolor{lightgray}{\scriptsize{[ICCV21]}}~\cite{son2021densely} & \reshx{74.39\%}{9.11\%} $\pm$ 0.37 \\
% ~~~~~~~~ -- w/ maximum schedule (${\rm{AMD}_{Max}}$)
\rowcolor{mygray}
${\rm{Swin-Base}_{10\%}}$ MMD
& \reshx{\textbf{76.86}\%}{6.64\%} $\pm$ 0.24 \\
\rowcolor{mygray}
% ~~~~~~~~ -- w/ minimum schedule (${\rm{AMD}_{Min}}$)
${\rm{Swin-Base}_{10\%}}$ AMD
& \reshx{\underline{76.81}\%}{6.69\%} $\pm$ 0.21 \\
\hline
\end{tabular}
}
\vspace{-0.4cm}
\end{wraptable}
\noindent \textbf{Results on ImageNet.} To further prove the effectiveness and generalization of our approach, we follow common practices~\cite{son2021densely, mirzadeh2020improved, liu2023norm}, and extend AMD into ImageNet~\cite{ImageNet}, presented in Table~\ref{table:main_results_imagenet}. Since we have shown in both CIFAR-10 and CIFAR-100 datasets that logit-based methods result in inferior performance on transformer-based architectures in Table~\ref{table:main_results}. We compare only feature-mimicking method CRD during this experiment on ViT and Swin. As seen, our method achieves over \textbf{2.61\%} better \texttt{top-1} accuracy than CRD on ViT-Base.

% subfigs ...
% \begin{figure}[t!]
% \begin{subfigure}{.5\textwidth}
%   \centering
%   % include first image
%   \includegraphics[width=1\linewidth]{imgs/efficient_1_TA.pdf}
%   \caption{\textbf{The sufficiency of using one teacher assistant.} \protect\includegraphics[scale=0.10,valign=c]{imgs/color_blocks.pdf} colors represent the performance of $\rm{AMD_{Min}}$ using zero, one, two, and three teacher assistants, respectively.}
%   \label{fig:sufficient}
% \end{subfigure}
% \begin{subfigure}{.5\textwidth}
%   \centering
%   % include second image
%   \includegraphics[width=1\linewidth]{imgs/epoch_efficient.pdf}
%   \caption{\textbf{Efficient training schedule.} Our proposed AMD enjoys superior performance among the overall training scene.}
%   \label{fig:epoch_efficient}
% \end{subfigure}
% \begin{subfigure}{.5\textwidth}
%   \centering
%   % include second image
%   \includegraphics[width=1\linewidth]{imgs/alpha_beta_update.pdf}
%   \caption{\textbf{Weight values from loss objectives.} We present the performance on different values of $\alpha$ and $\beta$. The highest performance is marked in \textcolor{red}{red} (\ie, $\alpha=0.2, \beta=100$). \protect\includegraphics[scale=0.10,valign=c]{imgs/alpha_beta_color_blocks.pdf} colors represent performance with respect to different $\beta \in \left\{1, 10, 50, 100\right\}$.}
%   \label{fig:alpha_beta}
% \end{subfigure}
% \end{figure}

\subsection{Diagnostic Experiments}\label{subsec:ablation}
We design several ablation studies to evaluate the robustness and effectiveness of our proposed AMD. Most experiments are conducted on ViT trained on CIFAR-100. For ablative study on the sufficiency of a single teacher-assistant, we report additional results on ViT trained on CIFAR-10 for completeness.
% delete base since new-added exp are from ViT-small

\setlength\intextsep{10pt}
\begin{wraptable}{r}{0.65\linewidth}
% \begin{table}[t]
\vspace{-0.7cm}
\caption{\textbf{Impact of student model scalability} ranging from 5\% to 20\%. In our study, we deliberately exclude models of a higher scale, taking into full account the constraints imposed by computational capacities.}
\vspace{-10pt}
\label{table:scalability}
\begin{center}
% \begin{small}
\tabcolsep=0.10cm
\resizebox{0.63\textwidth}{!}{
\begin{tabular}{c|c||r|c|c}
\hline \thickhline
\rowcolor{mygray}
Student scalability & Method & FLOPs  & Performance & GPU hours  \\
% \rowcolor{mygray}
% \rowcolor{mygray}
  % (86.7M)   &  &  \textit{Natural} [7] & \textit{Specialized} [4] & \textit{Structured} [8]   \\
\hline \hline
\multirow{2}{*}{$\mathrm{ViT-Base}_{5\%}$} & ADKD &\multirow{2}{*}{0.88G} & 68.44\% & $1\times$ \\
 & AMD & & \textbf{75.88\%} & $2.2\times$ \\
\hline
\multirow{2}{*}{$\mathrm{ViT-Base}_{10\%}$} & ADKD &\multirow{2}{*}{1.76G} & 73.69\% & $1\times$\\
& AMD & & \textbf{80.19\%} & $2.2\times$ \\
\hline
\multirow{2}{*}{$\mathrm{ViT-Base}_{15\%}$} & ADKD &\multirow{2}{*}{2.64G} & 75.17\% & $1\times$\\
& AMD & & \textbf{80.23} \% & $2.2\times$\\
\hline
\multirow{2}{*}{$\mathrm{ViT-Base}_{20\%}$} & ADKD & \multirow{2}{*}{3.52G} & 77.09\% & $1\times$\\
& AMD & & \textbf{81.09\%} & $2.2\times$\\
% \rowcolor{mygray}
% Ours &  0.21\% &  & \\
\hline
\end{tabular}
}
% \end{small}
\end{center}
\vspace{-2em}
% \end{table}
\end{wraptable}

\noindent \textbf{Impact of Student Model Scalability.} To investigate whether AMD can boost the performance across different student model scales, we report the performance of AMD with respect to scales ranging from 5\% of ViT-Base to 20\% in Table~\ref{table:scalability}. Specifically, when considering the 20\% ViT-Base with a computational demand of 3.52G FLOPs, it closely parallels the full-sized ResNet34 model, which requires 3.68G FLOPs with 77.94\%~\cite{binici2022preventing} \texttt{top-1} accuracy on CIFAR-100.
We intentionally refrain from expanding the student model to a larger scale since the primary objective of this paper is to maintain a compact student model. As the expansion of the student model in size leads to a notable increment in inference cost. This augmentation not only poses a significant challenge to the deployment flexibility but also intensifies the requisite computational resources. It can be seen that AMD achieves constantly strong performance with various model scales.
% We do not extend the student scale to a larger one since we focus on a compact student model, also, the training schdule would be extend when the scale continue to grow.

\begin{figure}[t!]
  \centering
       % \vspace{-16pt}
\includegraphics[width=0.90\textwidth]{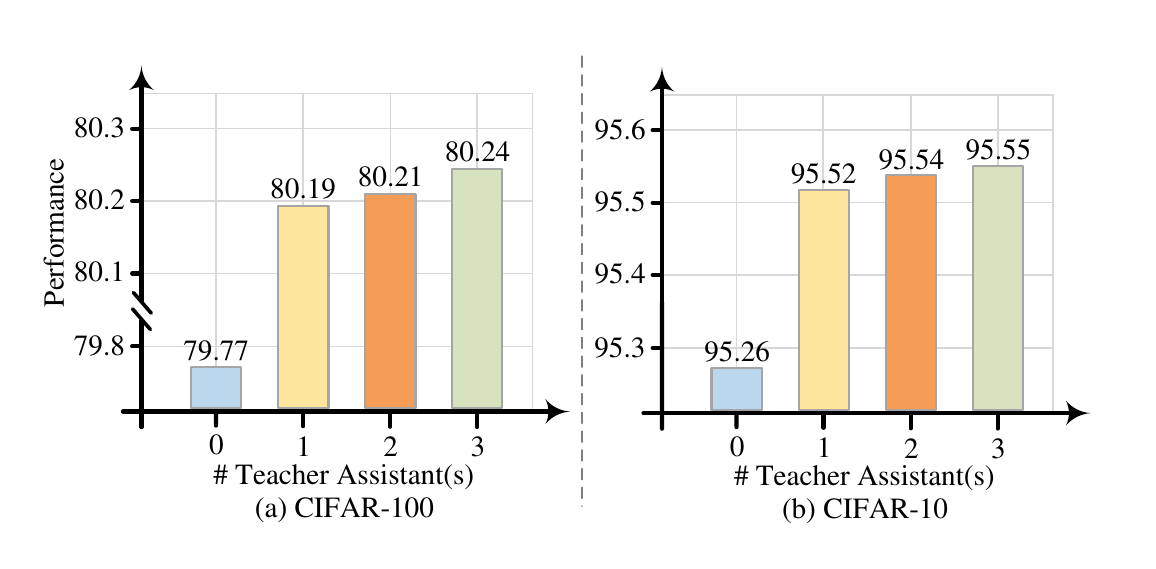}
\vspace{-20pt}
\caption{\textbf{The sufficiency of using one teacher-assistant.} \protect\includegraphics[scale=0.10,valign=c]{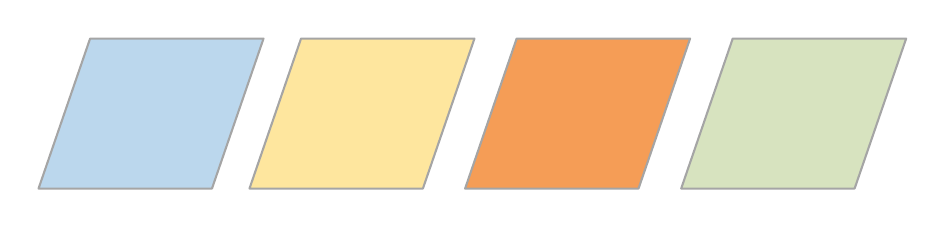} colors represent the performance of $\rm{AMD_{Min}}$ using zero, one, two, and three teacher-assistants, respectively. The results from the training on the ViT-Base CIFAR-100 is posited at (a), CIFAR-10 at (b).}
\label{fig:sufficient}
\vspace{-25pt}
\end{figure}

\noindent \textbf{Sufﬁciency of a Single Teacher-Assistant.}
To verify the efficiency of using one single teacher-assistant, we further set up ablation study on multiple teacher-assistants in our method. In Figure~\ref{fig:sufficient}\textcolor{red}{(a)}, we first demonstrate that the absence of the teacher-assistant results in a marked deterioration of performance (\ie, 79.77\%), which is consistent with the results from Table \ref{table:main_results}. We also find that the performance gain is marginal when having more than one teacher-assistant (\ie, 80.19\% $\rightarrow$ 80.24\%) with the dramatically increased GPU hours (\ie, $3\times$ and $4\times$ when having two- and three-step teacher-assistants, respectively).
The results obtained from the training on the ViT-Base CIFAR-10 are congruent with the observation (see Figure~\ref{fig:sufficient}\textcolor{red}{(b)}).
We therefore choose one teacher-assistant in AMD, enjoying both training efficiency and performance sufficiency. Additional results on both ViT and Swin are provided in the supplementary material.

% \begin{table}[t]
% % \vspace{-0.5cm}
% \caption{\textbf{Benefit on training schedule}}
% \vspace{-15pt}
% \label{table:training_schedule}
% \begin{center}
% \begin{small}
% \tabcolsep=0.10cm
% \resizebox{0.49\textwidth}{!}{
% \begin{tabular}{c||r|r|r|r|r}
% \hline \thickhline
% % \rowcolor{mygray}
% % Student scalability  & Epoch & Performance & GPU hours  \\
% % \rowcolor{mygray}
%   % (86.7M)   &  &  \textit{Natural} [7] & \textit{Specialized} [4] & \textit{Structured} [8]   \\
% % \hline \hline
% % $\mathrm{ViT-Base}_{5\%}$ & 0.88 \\
% % $\mathrm{ViT-Base}_{5\%}$ & \\
% % \hline
% \cellcolor{mygray} Epoch & 40 & 70 & 100 & 130 & 160 \\
% \hline
% % $\mathrm{ViT-Base}_{10\%}$ & 70 \\
% % \cellcolor{mygray} Performance & 79.51\% & 80.12\% & 81.44\% & 82.36\% & 82.53\% \\
% % \cellcolor{mygray} Performance & 70.82\% & 73.08\% & 78.75\% & 79.02\% & 79.17\% \\
% KD & 52.42\% & 55.74\% & 56.93\% & 57.67\% & 59.47\% \\
% DKD & 59.58\% & 64.72\% & 66.34\% & 68.13\% & 69.53\% \\
% TAKD & 60.37\% & 65.59\% & 68.67\% & 69.52\% & 70.89\% \\
% CRD & 65.98\% & 71.49\% & 74.65\% & 78.02\% & 78.29\% \\
% \cellcolor{mygray} $\rm{AMD_{Min}}$ & 70.82\% & 73.08\% & 79.15\% & 79.92\% & 80.19\% \\
% % \hline
% % $\mathrm{ViT-Base}_{15\%}$ & 2.64 \\
% % $\mathrm{ViT-Base}_{15\%}$ & \\
% % \hline
% % $\mathrm{ViT-Base}_{20\%}$ & 3.52 \\
% % $\mathrm{ViT-Base}_{20\%}$ & \\
% % \rowcolor{mygray}
% % Ours &  0.21\% &  & \\
% \hline
% \end{tabular}
% }
% \end{small}
% \end{center}
% \vspace{-1.5em}
% \end{table}

\begin{wrapfigure}{r}{0.55\textwidth}
% \begin{figure}[t!]
  \centering
       \vspace{-22pt}
\includegraphics[width=0.55\textwidth]{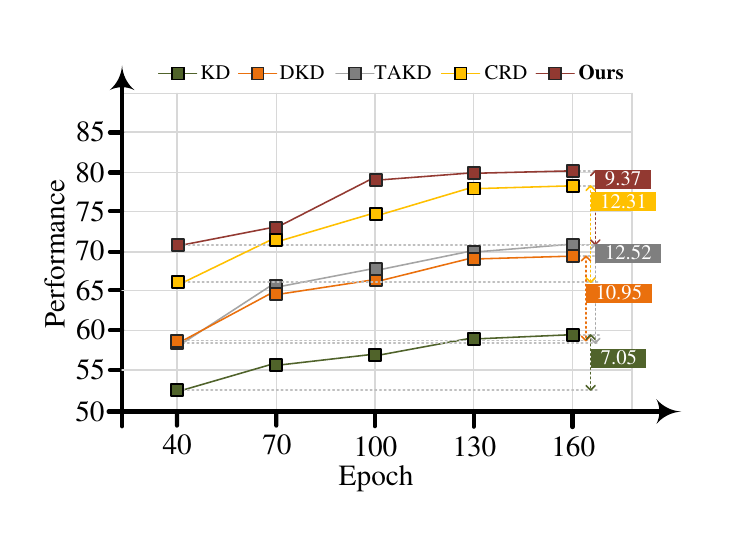}
\vspace{-30pt}
\caption{\textbf{Efficient training schedule.} Our proposed AMD enjoys superior performance among the overall training scene. }
\label{fig:epoch_efficient}
\vspace{-10pt}
% \end{figure}
\end{wrapfigure}

\noindent \textbf{Benefit on Training Schedule.}
Formally, we follow common practices (see \S\ref{subsec:experiment_setup}) and set 160 epochs as standard training schedule. However, we notice that AMD enjoys a fast convergence in early epochs.
% , mainly because the xxx (sandwich rule) enjoys weight sharing across different scales, thereby an extra initialization stage can be employed.
For example, in Figure~\ref{fig:epoch_efficient}, with only $40$ epochs, our AMD exhibits competitive performance (\ie, 70.82\% in CIFAR-100) in comparison to other prevalent methods (see Table~\ref{table:main_results}). Note that though KD~\cite{hinton2015distilling} exhibits narrow performance gap between epochs (\ie, 7.05\%), it can hardly reach a satisfying result under the setting of transformer-based architectures. AMD, on the other hand, performs consistently well even with a reduced number of epochs. This underscores the potential of our approach as an efficacious strategy for training in limited training schedule.

\vspace{-2pt}
\noindent \textbf{Impact of Candidate Sampling Rate.} We further study the variation of candidate sampling rate by changing the number of sampled candidates $m \in \left\{1,3,6,9,15\right\}$. A higher value of $m$ signifies a more refined granularity in the sampling rate. This increased
\setlength\intextsep{10pt}
\begin{wraptable}{r}{0.50\textwidth}
% \begin{table}[t]
\vspace{-0.3cm}
\caption{\textbf{Impact of candidate scaling $m$.}}
\vspace{-15pt}
\label{table:candidate_sampling}
\begin{center}
% \begin{small}
\tabcolsep=0.10cm
\resizebox{0.50\textwidth}{!}{
\begin{tabular}{l||c|c}
\hline \thickhline
\rowcolor{mygray}
Method & Performance &  GPU hours   \\
% \rowcolor{mygray}
% \rowcolor{mygray}
  % (86.7M)   &  &  \textit{Natural} [7] & \textit{Specialized} [4] & \textit{Structured} [8]   \\
\hline \hline
$\mathrm{ViT-Base}_{100\%}$ (teacher) & 89.33\% & - \\
~~~~~~ -- MMD   & 80.11\% & $20\times$\\
~~~~~~ -- AMD ($m=1$) & 78.39\% & $2\times$\\
~~~~~~ -- AMD ($m=3$) & 79.46\% & $2\times$\\
~~~~~~ -- AMD ($m=6$) & 79.84\% & $2.1\times$\\
~~~~~~ -- AMD ($m=9$) & 80.19\% & $2.2\times$\\
~~~~~~ -- AMD ($m=15$) & 80.22\% & $2.6\times$\\
% \rowcolor{mygray}
% Ours &  0.21\% &  & \\
\hline
\end{tabular}
}
% \end{small}
\end{center}
\vspace{-1.6em}
% \end{table}
\end{wraptable}
granularity is directly correlated with an extended duration of training time. The GPU hours and their corresponding student performance are reported in Table~\ref{table:candidate_sampling}. We set $m=9$ for a satisfying tradeoff between performance and computational overhead. An increased sampling rate invariably leads to a longer training time, which yields marginal enhancements in performance. For example, when having $m=15$, we observe $0.03\%$ performance gain can be achieved with $18\%$ GPU hour increment. We argue that this is inefficient for training schedule.

% \begin{table}[t]
% % \vspace{-0.5cm}
% \caption{\textbf{Impact of Different Loss Components}}
% \vspace{-15pt}
% \label{table:loss_component}
% \begin{center}
% \begin{small}
% \tabcolsep=0.10cm
% \resizebox{0.9\textwidth}{!}{
% \begin{tabular}{l||r}
% \hline \thickhline
% \rowcolor{mygray}
% Method &  Performance   \\
% % \rowcolor{mygray}
% % \rowcolor{mygray}
%   % (86.7M)   &  &  \textit{Natural} [7] & \textit{Specialized} [4] & \textit{Structured} [8]   \\
% \hline \hline
% AMD & 79.17\% \\
% ~~ -- w/o $\mathcal{L}_{CE}$ &  \\
% ~~ -- w/o $\mathcal{L}_{logit}$ & \\
% ~~ -- w/o $\mathcal{L}_{feat}$ & \\
% % \rowcolor{mygray}
% % Ours &  0.21\% &  & \\
% \hline
% \end{tabular}
% }
% \end{small}
% \end{center}
% \vspace{-1.5em}
% \end{table}

\begin{wrapfigure}{r}{0.60\textwidth}
% \begin{figure}[t!]
  \centering
       % \vspace{-16pt}
\includegraphics[width=0.60\textwidth]{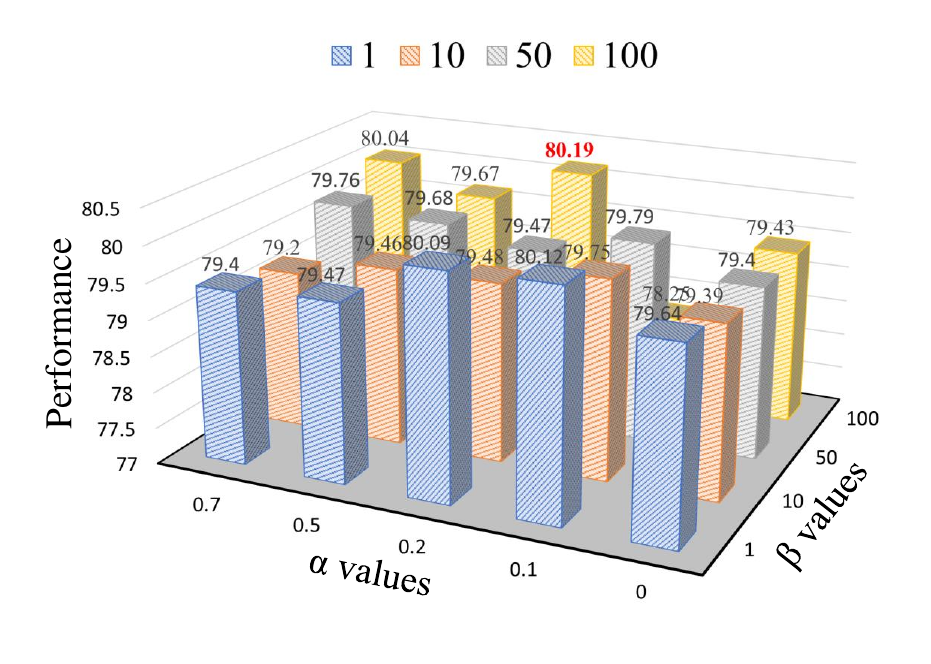}
% \vspace{-30pt}
\caption{\textbf{Weight values from different loss objectives.} We present the performance on different values of $\alpha$ and $\beta$ from Eq.~\ref{eq:overall_objective}. The highest performance is marked in \textcolor{red}{red} (\ie, $\alpha=0.2, \beta=100$). \protect\includegraphics[scale=0.10,valign=c]{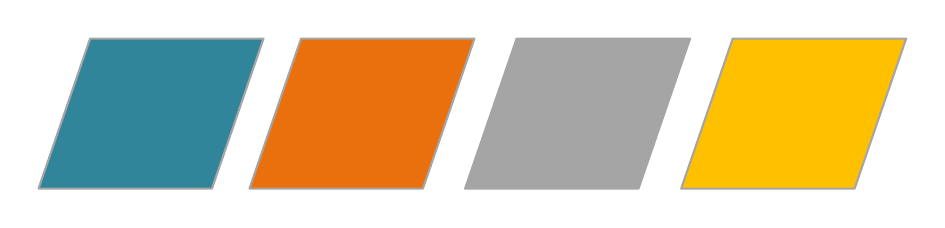} colors represent performance with respect to different $\beta \in \left\{1, 10, 50, 100\right\}$.}
\label{fig:alpha_beta}
\vspace{-15pt}
% \end{figure}
\end{wrapfigure}

\noindent \textbf{Impact of different Hyper-Parameters.} In Eq.~\ref{eq:overall_objective}, we introduce two hyper-parameters $\alpha$ and $\beta$ to balance the cross-entropy loss, logit-based loss and feature-mimicking based loss. We present the influence of different values of $\alpha$ and $\beta$ on performance in Figure~\ref{fig:alpha_beta}.
The results exhibit substantial robustness to the fluctuations in these hyper-parameters, evidenced by a low standard deviation of 0.42.
During experiment, a consistent pattern was discerned across various datasets: a larger $\alpha$ correlates with diminished influence over the teacher-assistant's logit responses, which in turn culminates in suboptimal performance results. Furthermore, we note a progressive enhancement in supervision efficacy on increasing the value of $\beta$. Unlike previous studies on NLP suggest a varying value of $\beta$ based on different tasks~\cite{wu2023ad}, $\beta$ is relatively stable across several datasets in our experiments.
% we observe a common trend across different datasets that when $\alpha$ is large, the contribution/supervision on teacher assistant's logit is limited, thereby result in a relatively low performance. Also, the supervision on xxx improved continuously until it reaches the peach and start to decline.
Overall, a hyper-parameter pairing of $\alpha=0.2$ and $\beta=100$ is attained for the best performance, which are the values used for all experiments.
% , showing that the results are relatively robust to hyper-parameters (\ie, with standard deviation 0.42) and get the highest performance with the combination of $\alpha=0.2, \beta=100$.

\noindent \textbf{Impact of Different Loss Components.} To analyze the impact of different loss components, we further conduct ablation studies on three variants of AMD: \textit{\textbf{\ding{182}}} AMD without cross-entropy loss $\mathcal{L}_{ce}$; \textit{\textbf{\ding{183}}} AMD without logit-based loss $\mathcal{L}_{logit}$; \textit{\textbf{\ding{184}}} AMD without feature-mimicking based $\mathcal{L}_{feat}$. As seen in Table~\ref{table:loss_component}, we observe a significant performance drop (80.19\% $\rightarrow$ 75.24\%) by removing the supervision on hidden states (\ie, $\mathcal{L}_{feat}$), which is consistent with our results in \S\ref{subsec:main_results} (\ie, performance are suboptimal via logit-based methods). The removal of $\mathcal{L}_{ce}$ and $\mathcal{L}_{logit}$ also cause a marked performance degradation (\ie, $80.19\% \rightarrow 78.32\%$ and $80.19\% \rightarrow 78.01\%$, respectively), underscoring the integral role both losses play in enhancing model efficacy.
\begin{wraptable}{r}{0.30\textwidth}
\begin{center}
\vspace{-0.9cm}
% \hspace{-1.0cm}
\caption{\textbf{Impact of different loss components}, including three variants from original training objectives (see Eq.~\ref{eq:overall_objective})}\label{table:loss_component}
% \vspace{-0.2cm}
\tabcolsep=0.10cm
% \hspace{-1.0cm}
\resizebox{0.30\textwidth}{!}{
\begin{tabular}{l||c}
\hline \thickhline
\rowcolor{mygray}
Method &  Performance   \\
% \rowcolor{mygray}
% \rowcolor{mygray}
  % (86.7M)   &  &  \textit{Natural} [7] & \textit{Specialized} [4] & \textit{Structured} [8]   \\
\hline \hline
% ${\rm{AMD}_{Min}}$
AMD
& 80.19\% \\
~~ -- w/o $\mathcal{L}_{CE}$ & 78.32\% \\
~~ -- w/o $\mathcal{L}_{logit}$ & 78.01\% \\
~~ -- w/o $\mathcal{L}_{feat}$ & 75.24\% \\
~~ -- w/ $\mathcal{L}_{dkd}$ & 80.22\% \\
\hline
\end{tabular}
}
\vspace{-0.7cm}
\end{center}
\end{wraptable}
Note that the influence of getting rid of $\mathcal{L}_{logit}$ has a higher impact on performance, which is consistent with our claim in the previous ablation study (\ie, $\alpha=1$).
% \textcolor{red}{
For completeness, we also conduct experiments on combining DKD~\cite{zhao2022decoupled} loss, which introduces target class knowledge distillation (\ie, TCKD) and non-target class knowledge distillation (\ie, NCKD) to further decompose $\mathcal{L}_{ce}$. Specifically, we have $\mathcal{L}_{dkd} = \zeta\mathrm{TCKD} + \eta\mathrm{NCKD}$, incorporating balancing parameters $\zeta$ and $\eta$. The result indicates that the DKD loss further improves the model performance (\ie, $80.19\% \rightarrow 80.22\%$). However, it is imperative to underscore that the incorporation of the DKD loss introduces additional hyper-parameters, which consequently engenders increased fluctuations in the quest to achieve optimal results. Therefore, we retain the original design as delineated in Eq.~\ref{eq:overall_objective} for the purpose of maintaining stability in the model's performance.
% , suggesting that both of them contribute performance improvements.

\begin{wraptable}{r}{0.30\textwidth}
\begin{center}
\vspace{-1.1cm}
% \hspace{-1.0cm}
\caption{\textbf{Discussion on NPSD metric.} We further introduce a general form of NPSD metric as $\lambda$-NPSD.}\label{table:npsd_metric}
\vspace{-0.1cm}
\tabcolsep=0.10cm
% \hspace{-1.0cm}
\resizebox{0.30\textwidth}{!}{
% \begin{adjustbox}{width=0.15\textwidth,right}
\begin{tabular}{l|| c }
    \hline \thickhline
        \rowcolor{mygray}
         & CIFAR-100 \\
        \rowcolor{mygray}
        \multirow{-2}{*}{Metric} & \texttt{top-1} \\
        \hline \hline
        NPSD ($\lambda=0$) & 80.19\% \\
        $\lambda=0.5$ & 80.19\% \\
        $\lambda=1$ & 80.14\% \\
                \hline
\end{tabular}
}
\vspace{-0.8cm}
\end{center}
\end{wraptable}
\noindent \textbf{Effectiveness of NSPD metric.}
% \textcolor{red}{
In order to prove the effectiveness of NSPD metric, we also include the combination of the negative derivatives between teacher and teacher-assistant, and teacher-assistant and student, \ie, $ - (\ \frac{P_{t} - P_{ta}}{S_{t} - S_{ta}} + \lambda \cdot \frac{P_{ta} - P_{s}}{S_{ta} - S_{s}})$, defined as a general form of NSPD metric (\ie, $\lambda$-NPSD). We report the results
% of ViT-Small with $\lambda=1$
in Table~\ref{table:npsd_metric} with different $\lambda$ values. It can be seen that $\lambda$-NPSD achieves on-par performance with original NPSD (we've tested other values of $\lambda$ as well). However, $\lambda$ introduces additional complexity, which also requires an additional distillation to obtain the performance of the student. Therefore, we choose to use the original NPSD from Eq.~\ref{eq:tradeoff}.

\vspace{-0.5mm}
\section{Conclusion and Discussion}\label{sec:Conclusion}

In contemporary vision research, a predominant focus has been on the knowledge transfer from convolutional neural networks. This emphasis tends to overlook the rapid advancements and proliferation of transformer-based architectures, both in terms of scale and accuracy. In light of this view, we propose AMD, a novel automatic multi-step distillation method for large-scale vision model compression. Unlike current knowledge distillation models in vision, AMD meticulously accounts for the scale-performance relation between the teacher and the teacher-assistant models. This consideration underpins the deployment of a cascade strategy that enables the elegant identification of the optimal teacher-assistant, ensuring a nuanced balance between training efficiency and performance efficacy. Comprehensive experiments demonstrate that AMD is able to: \textbf{I.} optimize training schedules for large-scale vision model distillation; \textbf{II.} ensure peak performance among competitive methods. We posit that our research constitutes a foundational contribution to knowledge distillation field, particularly as it pertains to large-scale vision models.
%Our findings have the potential to significantly advance the deployment of such models into computational constrained devices with uncompromising performance.
% We believe that our work make fundamental contributions to current knowledge distillation research on large-scale vision models.
% takes full consideration of the scale-performance relation between teacher and teacher assistant, thereby facilitating a cascade strategy to find the optimal teacher assistant elegantly.

\section*{Acknowledgement}
% We thank Prof. Majid Rabbani for detailed feedback on
% drafts of the paper.
This research was supported by the
National Science Foundation under Grant No. 2242243 and
the DEVCOM Army Research Laboratory under Contract
W911QX-21-D-0001.
% The views and conclusions contained
% herein are those of the authors and should not be interpreted
% as necessarily representing the official policies or endorsements, either expressed or implied, of the U.S. DEVCOM
% Army Research Laboratory (ARL), U.S. Naval Research
% Laboratory (NRL) or the U.S. Government.

% \section{Camera-Ready Manuscript Preparation}
% \label{sec:manuscript}
% This information will follow after paper decisions have been announced.

% \section{Conclusion}
% The paper ends with a conclusion.

% \clearpage\mbox{}Page \thepage\ of the manuscript.
% \clearpage\mbox{}Page \thepage\ of the manuscript.
% \clearpage\mbox{}Page \thepage\ of the manuscript.
% \clearpage\mbox{}Page \thepage\ of the manuscript.
% \clearpage\mbox{}Page \thepage\ of the manuscript. This is the last page.
% Now we have reached the maximum length of an ECCV \ECCVyear{} submission (excluding references).
% References should start immediately after the main text, but can continue past p.\ 14 if needed.
% \clearpage  % TODO REVIEW/FINAL: This \clearpage needs to be removed from both review and camera-ready versions.

\bibliographystyle{splncs04}
\bibliography{main}
\end{document}